\documentclass[letterpaper]{article} 
\usepackage{aaai22}  
\usepackage{times}  
\usepackage{helvet}  
\usepackage{courier}  
\usepackage[hyphens]{url}  
\usepackage{graphicx} 
\usepackage{natbib}  
\usepackage{caption} 
\usepackage{booktabs}
\DeclareCaptionStyle{ruled}{labelfont=normalfont,labelsep=colon,strut=off} 
\usepackage{algorithm}
\usepackage{algorithmic}

\usepackage{newfloat}
\usepackage{listings}
\floatstyle{ruled}
\newfloat{listing}{tb}{lst}{}
\floatname{listing}{Listing}
\title{On the Reliability of Large Language Models to Misinformed and Demographically-Informed Prompts}
\author{
    Toluwani Aremu\textsuperscript{\rm 1},
    Oluwakemi Akinwehinmi\textsuperscript{\rm 2},
    Chukwuemeka Nwagu\textsuperscript{\rm 3},
    Syed Ishtiaque Ahmed\textsuperscript{\rm 4},
    Rita Orji\textsuperscript{\rm 3},
    Pedro Arnau Del Amo\textsuperscript{\rm 2},
    Abdulmotaleb El Saddik\textsuperscript{\rm 1}\textsuperscript{\rm 5}
}
\affiliations{
    \textsuperscript{\rm 1} Mohamed Bin Zayed University of Artificial Intelligence, UAE,
    \textsuperscript{\rm 2} CIMNE, University of Lleida, Spain\\
    \textsuperscript{\rm 3} Dalhousie University, Canada,
    \textsuperscript{\rm 4} University of Toronto, Canada,
    \textsuperscript{\rm 5} University of Ottawa, Canada\\
    toluwani.aremu@mbzuai.ac.ae, aoluwakemi@cimne.upc.edu, cnwagu@dal.ca, ishtiaque@cs.toronto.edu, rita.orji@dal.ca, parnau@cimne.upc.edu, a.elsaddik@mbzuai.ac.ae, elsaddik@uottawa.ca

}

\usepackage{bibentry}
\usepackage{hyperref}

\begin{document}

\maketitle

\begin{abstract}
We investigate and observe the behaviour and performance of Large Language Model (LLM)-backed chatbots in addressing misinformed prompts and questions with demographic information within the domains of Climate Change and Mental Health. Through a combination of quantitative and qualitative methods, we assess the chatbots' ability to discern the veracity of statements, their adherence to facts, and the presence of bias or misinformation in their responses. Our quantitative analysis using True/False questions reveals that these chatbots can be relied on to give the right answers to these close-ended questions. However, the qualitative insights, gathered from domain experts, shows that there are still concerns regarding privacy, ethical implications, and the necessity for chatbots to direct users to professional services. We conclude that while these chatbots hold significant promise, their deployment in sensitive areas necessitates careful consideration, ethical oversight, and rigorous refinement to ensure they serve as a beneficial augmentation to human expertise rather than an autonomous solution.
\end{abstract}

\section{Introduction}

In recent times, the proliferation of Large Language Models (LLMs) has significantly impacted the field of artificial intelligence, owing to their exceptional capabilities in language comprehension and generation. These advanced models have become integral in various applications across multiple industries. Yet, their growing popularity and utility bring forth crucial challenges and ethical considerations.\\

\noindent Predominantly based on transformative deep learning architectures like Transformers, LLMs have revolutionized Natural Language Processing (NLP). These models, characterized by their vast neural networks containing millions or billions of parameters, are trained on extensive datasets encompassing a wide array of sources such as internet content, literary works, and diverse media. Such comprehensive training enables them to grasp and interpret a myriad of linguistic patterns and subtleties.\\

\noindent Mirroring the historical reliance on search engines for internet queries, users are now increasingly turning to chatbots powered by LLMs for instantaneous and direct responses. Notably, since the advent of ChatGPT, a variant based on the GPT-3.5 architecture in late 2022, the development and implementation of LLMs have rapidly expanded across various sectors. These models have been deployed in areas including virtual assistance, customer support, content creation, search functionality, and in the realms of medical, scientific research, programming assistance, educational tools, and more. However, this expansion has simultaneously sparked significant concerns regarding the ethical use of these technologies, as there have been instances of the models exhibiting biases, generating inaccurate information, making unfair judgments, or inciting ethical debates due to potential misuse.\\


\noindent Large Language models (LLMs) are repidly evolving, raising concerns about their potential to generate and disseminate misinformation. Biases within these models could lead to unequal information access or reinforcement of existing societal biases. Based on these issues, we investigate the behaviour of chatbots backed by LLMs, to answer two research questions;

\subsection{Research Questions} 
\begin{enumerate}
    \item When faced with misinformed prompts, do LLMs reflect, amplify, or rectify the misinformation through their responses?
    \item Do LLMs exhibit biases when answering prompts which contain demographic information?
\end{enumerate}

\noindent To answer these questions, we focused on the implications of utilizing these chatbots in discussions related to climate change and mental health. We focus our analysis on three LLM-powered chatbots: ChatGPT, Bing Chat, and Google BARD, assessing whether they manifest biases or propagate misinformation. Climate change and mental health, being among the most extensively discussed topics on social media as indicated by Google Trends\footnote{\url{https://trends.google.com/trends/}} and Exploding Topics\footnote{\url{https://explodingtopics.com/}}, are chosen for their relevance and the critical nature of accurate information dissemination in these areas. For the purpose of our study, our main contributions are as follows:
\begin{itemize}
    \item We developed a comprehensive benchmark dataset comprising 3,120 true/false questions on climate change and 2,762 on mental health. This dataset was instrumental for the empirical and quantitative evaluation of responses from LLM-backed chatbots.
    
    \item We conducted an in-depth qualitative analysis in collaboration with domain experts to scrutinize the responses from ChatGPT, Google BARD, and Bing Chat for potential biases. The findings from this analysis are presented herein. To support this, we constructed a dedicated benchmark dataset containing 53 questions on climate change and 40 on mental health, aimed at analyzing the extent of misinformation in the responses provided by these chatbots. Additionally, we utilized 24 climate change and 38 mental health questions specifically to evaluate biases.
    
    \item We proposed and deliberated on several strategies that could address the current challenges hindering the effective and ethical deployment of LLM-backed chatbots in providing accurate information on climate change and mental health issues.
\end{itemize}

\section{Literature Review}
\subsection{Background}
The contemporary AI landscape, particularly the rise of large language models (LLMs) has sparked critical discussions around ethical concerns. These concerns extend beyond job displacement and privacy violations to encompass the potential for misinformation dissemination. LLMs, trained on massive datasets, can unknowingly perpetuate biases and factual inaccuracies present in the training data \cite{46_bias}. Previous studies says confirmation bias and motivated reasoning can lead to favor information that aligns with existing beliefs \cite{47_confirmation_bias}. This raises concerns about the trustworthiness of LLMs outputs, especially when applied to sensitive domains like climate change and mental health. \\

\noindent Massive foundation models, which have found applications across a wide array of domains and contexts \cite{2_foundation_models_survey}. These models, boasting billions of learned parameters and trained on extensive datasets, have exhibited remarkable effectiveness in their respective downstream tasks. Consequently, the integration of AI into real-world applications has witnessed a phenomenal and exponential surge \cite{1_foundation_models_multimodal, 3_foundation_models_medical, 4_generative_ai}. In sharp contrast to traditional models, which often suffer from inherent constraints tied to their narrow focus, foundation models offer a versatile and adaptable approach. Once these models have undergone training, they can be conveniently fine-tuned to suit a diverse spectrum of applications, thus eliminating the necessity for extensive retraining. This adaptive framework serves as the linchpin of LLMs. Notably, this fine-tuning capability has paved the way for deploying these language models in a multitude of domains, spanning healthcare, financial advisory, climate change analysis, and question answering, to name just a few. \\

\noindent As the scope of artificial intelligence (AI) continues to expand, ushering in captivating innovations, it has sparked spirited debates on a broad range of ethical concerns. These concerns encompass the potential impacts of these advancements on various facets of human existence, including individual lives, employment, privacy, and issues related to discrimination \cite{5_bias_employment, 6_perception}. Simultaneously, questions have emerged regarding the appropriate course of action for the adoption of these transformative technologies \cite{7_pandora, 8_bias}. \\

\noindent According to an article by researchers at Google published in 2021 \cite{6_perception} to assess public perception of AI in eight countries, people in developing countries like Nigeria, India, and Brazil are significantly more likely to embrace and adopt AI compared to individuals in developed countries. However, ethical researchers in the AI domain have raised concerns that such AI applications may disproportionately affect people living in these regions, as most of the data used to train these models is sourced from developed countries \cite{9_gebru, 10_sch, 11_buolamwini, 12_ini, 13_ini, 14_brent}. Therefore, it comes as no surprise that throughout 2023, a series of pivotal governmental hearings have convened, where political leaders engaged with a diverse array of experts to gain insights into the origins and implications of these technologies. These hearings have probed critical aspects, such as the inherent risks associated with these AI systems, the nature of the data on which they are trained, and the formulation of policies designed to safeguard the well-being and privacy of users. These discussions also consider equitable compensation for the creators and owners of the data that underpin these AI systems. \\

\noindent In this section, our focus narrows to articles highlighting the deployment of LLMs in the contexts of climate change/sustainability and mental health/physical health. Our objective here is to demonstrate the substantial strides made in the adoption of AI technologies, setting the stage for subsequent sections where we delve into our methodologies and present the results of experiments conducted to evaluate biases and misinformation in LLMs when applied to both climate change and mental health contexts.

\subsection{Climate Change}
The advent of large language models in the realm of climate change research took a significant leap forward in 2021 with the introduction of ClimateBERT by Webersinke \cite{15_webersinke}. ClimateBERT, a transformer-based language model, was pretrained on an extensive dataset comprising over 2 million paragraphs sourced from climate-related texts, including news, research articles, and corporate climate reports. Its primary purpose was to facilitate climate change question answering and text summarization.  Subsequently, an array of tools \cite{16_chatclimate,17_climatechat,18_climatchat,19_climatechat,20_climatechat,21_climatechat} and datasets \cite{22_climatedata,23_climatedata,24_climatedata} have been created to improve the credibility and correctness of information disseminated by applications utilizing such models.\\

\noindent While several empirical studies have examined the effectiveness of these tools, most have concentrated on sentiments \cite{25_climatesentiment,26_climatesentiment,27_climatesentiment,28_climatesentiment} and sustainability \cite{29_climatesustainability}. In a closely related study \cite{30.climateassessing}, researchers assessed the accuracy of LLMs in handling climate information and proposed a practical protocol that combines AI assistance with human raters to mitigate the limitations encountered during the evaluation of LLMs. Our work, in contrast, encompasses both qualitative and quantitative evaluations of LLM responses to misinformed queries and considers how these models interact with various demographic groups.

\subsection{Mental Health}
Language models show promise in addressing challenges within the field of mental health. Several of these models, employing smaller language models \cite{31_mental, 32_mental}, have been proposed for applications in both mental health and general medical care. More recent developments have leveraged larger language models \cite{33_mental, 34_mental, 35_mental, 36_mental}. It's important to note that these models introduce ethical concerns, as they have the potential to cause irreversible harm to users.\\

\noindent Empirical evaluations of LLMs in this domain primarily fall into two categories: some assess LLMs' ability to classify different types of mental health issues using annotated text data \cite{37_mental, 38_mental}, while others evaluate their performance in various medical examinations \cite{39_mental, 40_mental, 41_mental, 42_mental, 43_mental, 44_mental}. Our approach, however, diverges from these studies. We delve deeper into the analysis of the responses generated by the models we employ, collaborating with experts to determine their readiness for real-world deployment or whether significant strides are still required in this area.

\section{Methodology}
This study is structured to address our central research objective which is to evaluate the level of misinformation and bias in LLM-powered chatbots in climate change and mental health discussions. We do this through a dual-pronged approach: firstly, by understanding and quantifying misinformation, and secondly, by evaluating biases in the responses of LLM chatbots. This section details the methodologies employed in each of these categories.

\subsection{Tools and Data Collection}
In our study, we evaluate three cutting-edge most popular and accessible LLM chatbots: Microsoft's Bing Chat, OpenAI's ChatGPT, and Google's Bard (now Gemini). To conduct an extensive evaluation, we compiled a set of frequently asked questions (FAQs) on two crucial topics—Climate Change and Mental Health. We prioritize frequently asked questions (FAQs) to reflect real-world user queries encountered by these LLMs. This approach ensures the generalizability of our findings to real-world LLMs interactions. We also assume that the LLMs used in these chatbots were trained on and has access to the information. Hence, to test each chatbot susceptibility to misinformation, we intentionally altered a subset of the selected questions-.These alterations involved introducing subtle factual inaccuracies, changing the tone of the prompt, or incorporating irrelevant words to the prompts. The specific type of alteration depended on the misinformation concept we aimed to assess. We used the same misinformed prompts uniformly across the chatbots intended for testing. 

\paragraph{Misinformation Assessment:}
For the quantitative analysis of misinformation, we curated a dataset comprising 3,120 true/false questions on Climate Change and 2,762 on Mental Health. For the qualitative analysis, we selected 53 questions about Climate Change and 40 about Mental Health from authoritative sources like NASA and the CDC. These prompts were intentionally altered to test if these chatbots are capable of handling the misinformed prompts, or would rather amplify the misinformation.

\paragraph{Bias Assessment:}
In evaluating potential biases in the chatbots' responses, we chose 24 questions on Climate Change and 38 on Mental Health. The aim here was to conduct a qualitative assessment of the chatbots' outputs, checking for objectivity and neutrality. To this end, specific demographic details such as age, race, and location were integrated into the prompts.

\paragraph{Chatbot Query Approach:}
For Qualitative Analysis, we interacted with the chatbots using a standardized format to ensure concise and informative responses. Each question was formatted as:
\begin{center}
\texttt{"In one short paragraph, \textbf{[question]}. Provide sources for your response."}
\end{center}

\noindent For Quantitative Analysis, the methodology for quantitative interaction also employed a standard prompt format, simplifying the chatbots' responses to a binary choice:
\begin{center}
\texttt{"Respond with either True/Yes or False/No: \textbf{[question]}."}
\end{center}

\subsection{Analysis}
This section details the methodological framework adopted for analyzing the misinformation and bias components in our study, divided into two distinct segments:

\paragraph{Quantitative Analysis to check for Misinformation:}
The quantitative analysis scrutinizes the chatbots' responses using a suite of metrics:
\begin{itemize}
    \item \textit{Confusion Matrix:} This tool visualizes the distribution of True Positives, False Positives, True Negatives, and False Negatives for the true/false questions.
    \item \textit{Precision and Recall:} These metrics evaluate the accuracy and completeness of the classification model, based on the results of the confusion matrix.
    \item \textit{F1 Score and Accuracy:} These indicators provide insights into the model's harmonic balance between Precision and Recall.
    \item \textit{Similarity Index Scores:} We employ BLEU \cite{papineni-etal-2002-bleu}, ROGUE \cite{lin-2004-rouge}, and METEOR \cite{banerjee-lavie-2005-meteor} scores to measure the closeness of the chatbot's responses to standard benchmark answers, thereby assessing the quality and relevance of the content provided.
\end{itemize}

\paragraph{Qualitative Analysis to check for Misinformation:}
We conduct a comprehensive qualitative analysis of the responses from the chatbots on the dataset we collected for this analysis. This facet of the analysis involves in-depth interviews with domain experts and the deployment of specialized questionnaires tailored to these fields.

\paragraph{Qualitative Analysis to check for Bias:}
The bias analysis segment focuses on quantitatively evaluating feedback from domain experts. These specialists will critique and provide perspectives on the extent of bias evident in the chatbots' responses. The aim here is to uncover any subjective biases that might be embedded in the outputs related to Mental Health and Climate Change topics.

\begin{figure}[t]
\centering
\includegraphics[width=\linewidth]{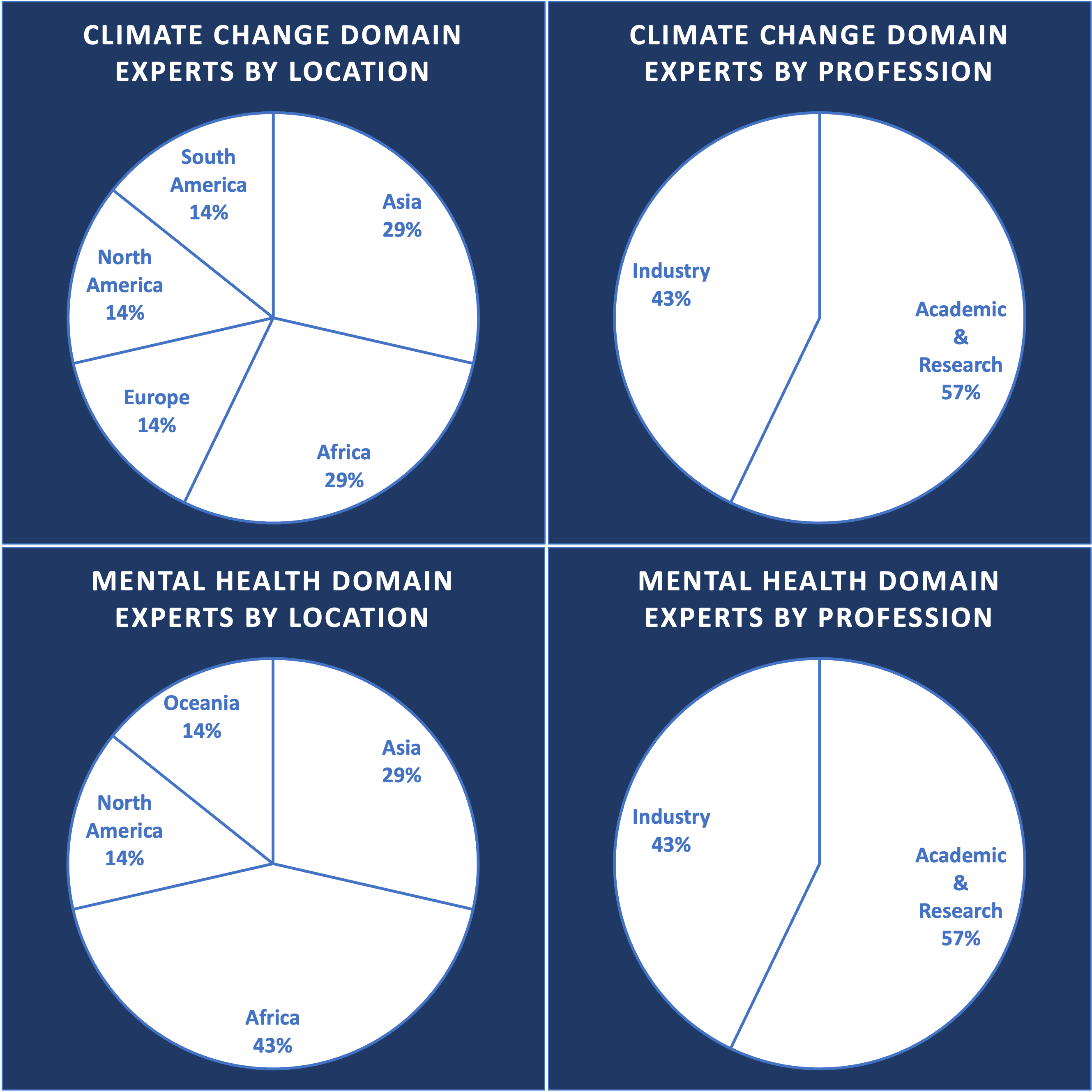}
\caption{Distribution of domain experts in the fields of Climate Change and Mental Health, categorized by location and profession. In both domains, Asia and Africa provide the largest regional share of experts, while North America and Oceania contribute the least. The majority of experts in both domains are from Academic \& Research backgrounds, accounting for 57\% of the total, as opposed to 43\% from Industry.}
\label{fig:experts-distribution}
\end{figure}

\begin{figure*}[t]
\centering
\includegraphics[width=\textwidth]{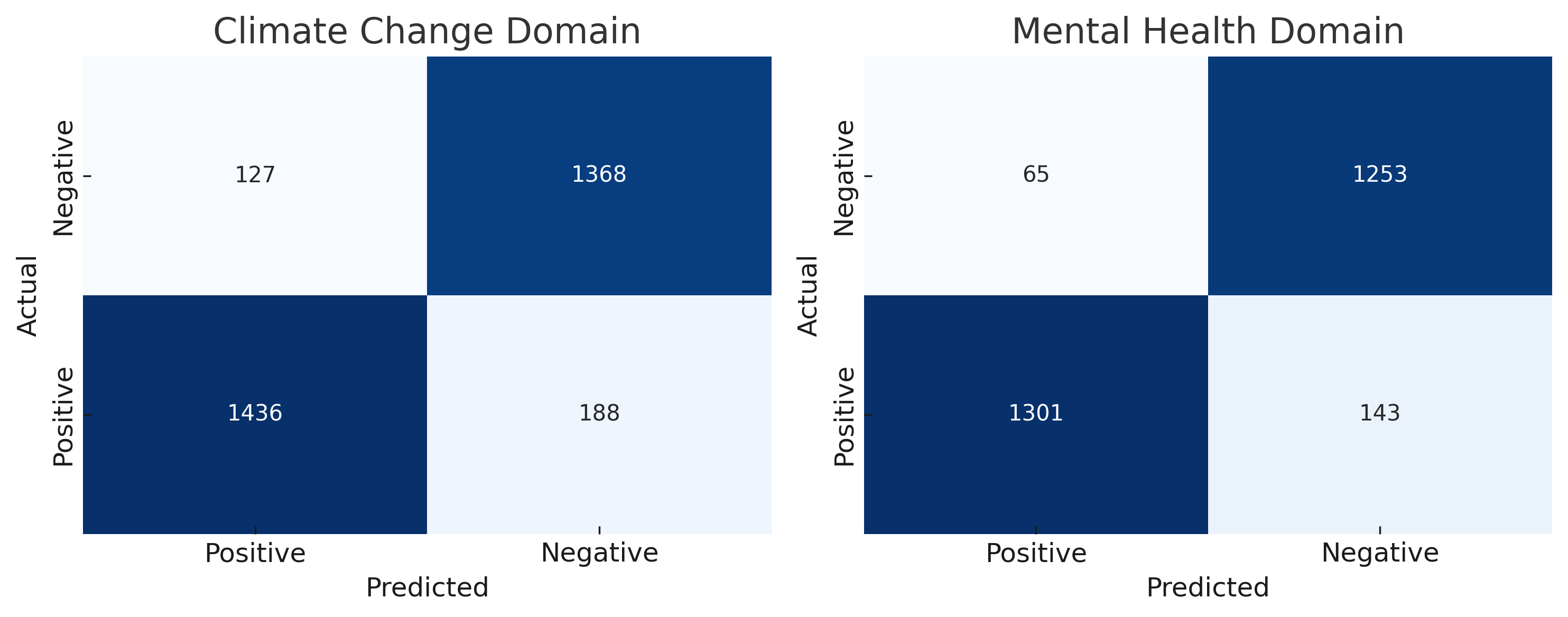}
\caption{Confusion matrices depicting the performance of the selected chatbots in answering whether a fact given within a prompt is either true or false, for the Climate Change and Mental Health domains. For Climate Change, there were 1,368 true negatives and 1,436 true positives, with false positives and negatives at 188 and 127, respectively. In the Mental Health domain, the model produced 1,253 true negatives and 1,301 true positives, and lower false positives and negatives at 143 and 65. These results indicate a high level of accuracy in the model's knowledge across both domains.}
\label{fig:confusion-matrices-corrected}
\end{figure*}

\subsection{Domain Expert Selection Criteria}
In the process of selecting domain experts for our study, we established specific criteria tailored to the distinct fields of Climate Change and Mental Health. \\

\noindent For Climate Change, we targeted academically credentialed professionals, including Professors, PostDocs, PhD students, researchers, or practitioners holding at least a master's degree in fields such as environmental science, climatology, meteorology, or ecology. Their expertise was validated through a demonstrated track record in climate change research, including publications in peer-reviewed journals, conference presentations, or significant contributions to relevant industry projects. A prerequisite was a minimum of three years of active involvement in areas such as climate change research, policy development, mitigation strategies, adaptation methods, or advocacy. Additionally, we emphasized the importance of interdisciplinary knowledge, combining insights from atmospheric science, oceanography, and social sciences, to foster a comprehensive understanding of climate change impacts. Familiarity with climate policies, the ability to effectively communicate complex scientific concepts, and experience in innovative solutions and collaborative projects were also deemed essential.\\

\noindent In the Mental Health domain, our focus was on professionals with a solid educational foundation in psychology, psychiatry, clinical social work, counseling, or related disciplines, requiring a minimum of a master's degree. We sought experts with substantial clinical or research experience in mental health, evidenced by a history of patient care, participation in clinical trials, research contributions, or advocacy work. Proficiency in various therapeutic modalities such as cognitive-behavioral therapy, psychotherapy, and mindfulness-based interventions was crucial. Cultural competence—understanding and addressing the diverse cultural and socioeconomic factors influencing mental health—was another critical criterion. Lastly, we valued experts open to exploring the ethical implications and potential applications of generative language technologies in mental health care and challenges.

\subsection{Limitations}
\label{sec:limitations}
A significant challenge encountered was the recruitment of domain experts. For interview-based qualitative reviews, standard practice recommends a minimum of five experts per domain. Questionnaire-based qualitative reviews generally require a more extensive participant base, ideally with at least 50 respondents. Despite extensive outreach efforts, our response rate was limited to 14 participants, comprising 4 interviewees and 10 questionnaire respondents. This equates to seven domain experts for each of the two categories under study. Although the number of participants are limited, the use of both quantitative and qualitative approach offer opportunity for indepth data and insights. We also believe that the inclusion of experts from diverse backgrounds, culture, and continents contributes positively to the quality of our findings.\\

\noindent As mentioned earlier, one of the strengths of this study is the geographical diversity of our expert panel. We successfully included at least one domain expert from each continent (refer to Figure \ref{fig:experts-distribution} for details), which bolsters the validity of our results. This diverse representation helps mitigate regional biases and enhances the global relevance of our findings.\\

\noindent Hence, we believe that despite the limitation in terms of numbers, the study provides meaningful insights into the research area. The limitations highlight avenues for future research, particularly in broadening the expert participant base to further validate and enrich the study's conclusions.

\section{Findings}
This section elucidates our study's results, commencing with a quantitative analysis of the chatbots' average performance on the dataset of True/False questions within the Climate Change and Mental Health domains. We subsequently present the similarity scores based on the metrics mentioned above, comparing the responses of all three chatbots against established facts to determine their factual adherence. Lastly, we expand into the qualitative insights derived from engaging with domain experts, providing an in-depth exploration of their perspectives in both domains of interest. These findings are consistent with research by \cite{4_generative_ai, 5_bias_employment, 10_sch, 46_bias, 47_confirmation_bias} who found that LLMs trained on massive datasets can still be susceptible to misinformation, particularly when the information is cleverly disguised.

\begin{table}[t]
\centering
\caption{Comparative analysis of the selected chatbots' performance in discerning True/False statements within the realms of Climate Change and Mental Health.}
\resizebox{\linewidth}{!}{%
\begin{tabular}{lcccc}
\toprule
\textbf{Domain} & \textbf{Precision} & \textbf{Recall} & \textbf{F1 Score} & \textbf{Accuracy} \\
\midrule
Climate Change & 88.4\% & 91.9\% & 90.1\% & 89.9\% \\
Mental Health  & 90.1\% & 95.2\% & 92.6\% & 92.5\% \\
\bottomrule
\end{tabular}%
}
\label{tab:performance_metrics}
\end{table}

\subsection{Quantitative Analysis: True/False Prompts}
In this study, we evaluate the chatbots' ability to discern the veracity of statements related to climate change and mental health, in order to quantify its level of knowledge, or how misinformed it might be. Utilizing a quantitative approach, we analyze a True/False dataset and calculate critical performance metrics. The analysis (Figure \ref{fig:confusion-matrices-corrected}) includes a detailed examination of instances where the model incorrectly classified true statements as false (false negatives) and false statements as true (false positives), as well as accurately identified true (true positives) and false (true negatives) statements.\\

\noindent An in-depth analysis of the performance metrics, as detailed in Table \ref{tab:performance_metrics}, shows that in the realm of Climate Change, our chatbots demonstrates commendable accuracy with a precision rate of 88.4\%. This indicates that the majority of the statements classified as true by the model are indeed correct. The recall rate of 91.9\% further suggests that it successfully identifies a high percentage of the true statements within this domain. The F1 score stands at 90.1\%, reflecting a strong overall performance. However, the overall accuracy, at 89.9\%, while high, indicates there is room for improvement in reducing misinformation when it comes to climate change. \\

\noindent In the Mental Health domain, the observed performance is notably enhanced. It achieves a higher precision rate of 90.1\%, suggesting that its capacity to correctly identify true statements is more refined in this domain. The recall rate of 95.2\% is particularly impressive, indicating that the model is highly effective at capturing true instances. The F1 score, at an elevated 92.6\%, points to a balanced and efficient classification capability. Moreover, the accuracy of 92.5\% underscores a significant level of reliability in the Mental Health domain.

\begin{figure}[ht]
\centering
\includegraphics[width=\linewidth]{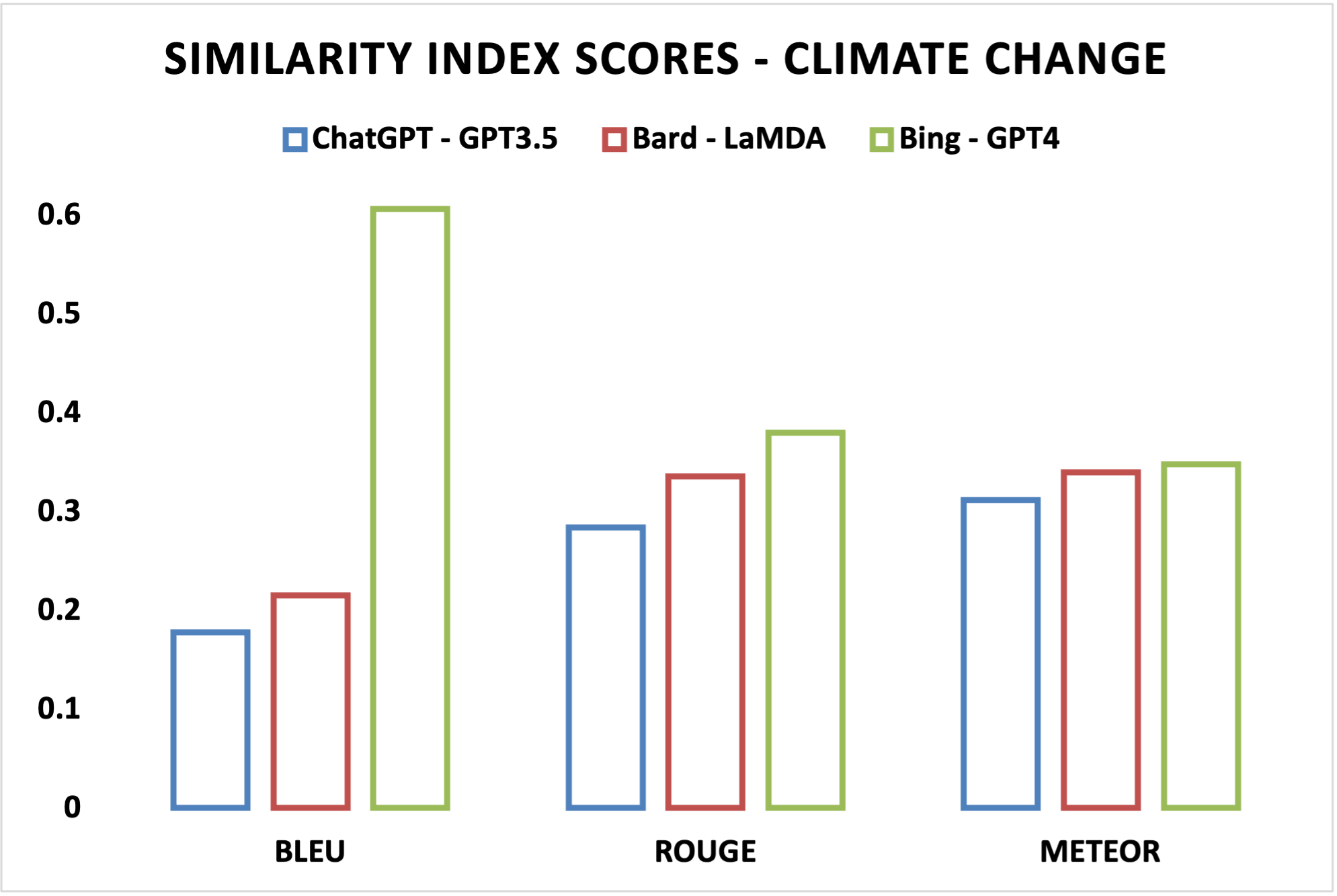}
\includegraphics[width=\linewidth]{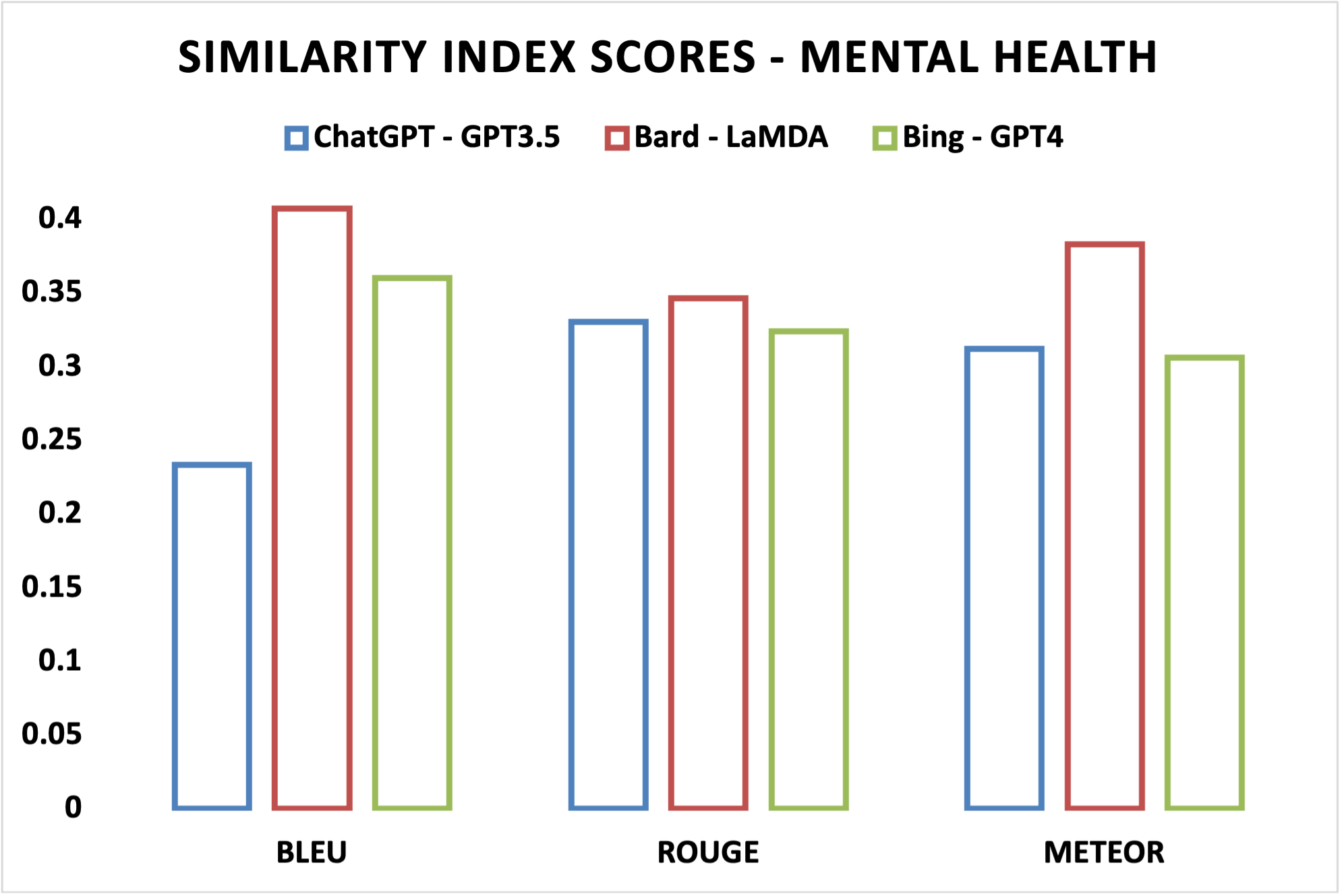}
\caption{The bar charts illustrate the Similarity Index Scores for three LLM-powered chatbots—ChatGPT (GPT-3.5), Bard (LaMDA), and Bing (GPT-4)—across three evaluation metrics: BLEU, ROUGE, and METEOR.}
\label{fig:similarity_index_scores}
\end{figure}

\subsection{Quantitative Analysis: Similarity Index Scores}
In the quantitative phase of our analysis, we employed three well-known metrics—BLEU, ROGUE, and METEOR—from the machine translation evaluation field. These metrics traditionally assess how closely machine-generated text matches human translation, in terms of both accuracy and contextual coherence. For this study, we adapted these metrics to assess the performance of chatbots, positing their applicability beyond their usual context of translation.\\

\noindent BLEU and ROGUE metrics are designed to measure the precision and recall of the chatbots' responses against a benchmark of human-generated texts. METEOR goes a step further by including advanced linguistic analysis—such as synonym matching, stemming, and paraphrasing—to provide a more nuanced assessment. This metric therefore offers a measure of evaluation that more closely approximates human judgment by accounting for semantic and contextual accuracy, in addition to exact word correspondences. We analyzed the chatbots' outputs by comparing them to the verifiable answers within our dataset. To ensure comparability, we normalized the resulting similarity scores, aiming for a maximum value of 1. This step was crucial, given that our `misinformed' prompts often led to chatbot responses that were shorter and substantially varied from the factual responses, sometimes resulting in inaccuracies or misinformation. Although there's no absolute threshold set for misinformation, these normalized scores serve as indicators of the degree to which the chatbots' responses emulate the factual data.\\

\noindent Figure \ref{fig:similarity_index_scores} presents the normalized similarity index scores, which reflect the chatbots' accuracy in relation to the factual statements provided. Within the climate change context, the Bing Chatbot, powered by GPT-4, demonstrated the highest likelihood of delivering correct responses—even when faced with misleading prompts. Google's Bard, leveraging LaMDA, followed closely, likely benefiting from its access to up-to-date online information, in contrast to ChatGPT (GPT-3.5), which operates based on data up until 2021 and hence offline. In the mental health arena, Bard was observed to have the highest accuracy, suggesting that its model is well-equipped to handle even adversarially designed prompts.

\subsection{Qualitative Analysis}
\label{sec:qualitative_analysis}
This section details our approach to gauging the levels of misinformation and bias present in the responses provided by three prominent chatbots—specifically, those focused on Climate Change and Mental Health topics. To this end, we sought insights from domain experts in these respective fields.\\

\noindent Initially, we endeavored to engage a broad spectrum of specialists for in-depth interviews based on the chatbots' responses. As revealed in Section \ref{sec:limitations}, the response rate was limited to 14 experts, spanning both domains. Our initial strategy involved forwarding the chatbot-generated responses to these experts, followed by interviews after a week. However, time constraints necessitated a strategic switch after interviews with our first two climate change experts. We transitioned to a questionnaire-based approach, using similar questions from the interview methodology, which significantly enhanced time efficiency. \\

\noindent The questionnaires comprised both closed- and open-ended questions, adapted to suit the experts' convenience. This approach was similarly applied in the mental health domain, where two specialists were interviewed, and the remaining provided their inputs via questionnaires. The details of the dataset used, chatbots' responses, and assessment questions are accessible here.\footnote{Data: \url{https://shorturl.at/hDHY0}}\\

\noindent The expert interviews varied from 45 to 60 minutes, encompassing 8 to 13 comprehensive questions, indicative of the depth of these discussions. Conversely, the questionnaires, comprising 11 items for climate change experts and 16 for mental health experts, required 10 to 30 minutes to complete.\\

\noindent The climate change-focused questions sought expert opinions on Language Model-based chatbots in raising awareness and their effectiveness in climate change communication. The assessment covered diverse aspects, including their role in awareness, challenges in providing accurate information, potential biases and misinformation, their utility in adaptation and mitigation strategies, user engagement features, integration with other platforms, promoting sustainable behaviors, and the relevance of their information in light of new scientific discoveries.\\

\noindent In the mental health context, the questions evaluated the impact, ethical considerations, effectiveness, and limitations of chatbots. Key areas of inquiry included their role in stigma reduction and awareness, personalized support effectiveness, specific mental health conditions addressed, necessity for human intervention, fostering trust and confidentiality, early detection and prevention, inclusivity, cultural sensitivity, potential drawbacks, criteria for success measurement, complementing existing services, empathy level, and their ability to recommend tailored mental health resources.\\

\noindent Our analysis of expert responses was conducted using a thematic analysis approach, adhering to the guidelines outlined by Braun and Clarke \cite{45_Braun_Clarke_2006}. This method allowed for the systematic identification and organization of visible and valid patterns within the data. Initially, we iteratively read through the responses to extract significant statements and concepts concerning the use and implications of chatbots in the domains of Climate Change and Mental Health, which we then represented as codes. Thematic saturation was achieved when no new codes could be identified. In the final phase of our analysis, we synthesized these themes into a coherent narrative. This involved linking the themes to our research questions and drawing conclusions about the role and impact of chatbots in the respective fields. Our interpretations, grounded in the data, include representative quotes from the experts to illustrate their perspectives.\\

\noindent The subsequent sections of this paper will delve into these specific themes, providing a detailed exploration of the utility and effectiveness of chatbots in the context of Climate Change and Mental Health.

\paragraph{(I.) Experts' Perspectives on Misinformation and Bias in Climate Change:}
In our investigation, we presented each expert with questions concerning the chatbots' responses on Climate Change. The focus of these inquiries was to understand the experts' perceptions of the potential impact these chatbots might have on user safety and information dissemination. We present our findings under four main themes.

\begin{enumerate}
    \item \textbf{Role of LLM-based Chatbots in Climate Change Awareness:} We asked the experts about the significance of LLM-based chatbots in enhancing public awareness of climate change. The majority, barring two, were optimistic, acknowledging that the newer generation of chatbots could play a substantial role in spreading awareness and disseminating information. On the contrary, one expert expressed skepticism about the depth and utility of the chatbots' responses, likening them to shallow internet searches. Another pointed out no discernible advantage over traditional search engines. Despite these differing views, there was a consensus that more specialized and expert-driven models could yield more precise and reliable information than the three chatbots evaluated. The experts suggested improvements such as enabling chatbots to provide simplified yet comprehensive answers, elucidate the reasoning behind their responses, and transparently cite their information sources.
    
    \item \textbf{Challenges in Generating Accurate Climate-Related Information:} We sought the experts' views on the obstacles faced by chatbots in delivering precise and trustworthy information on climate change. A significant portion of the respondents raised concerns about the data sources used to train these models. They emphasized the lack of assurance regarding the quality and credibility of the sources cited by the chatbots. Another common issue highlighted was the inconsistency in the chatbots' responses. Experts noted that for certain queries, the models produced vastly differing answers, which could lead to confusion. Additionally, the generality of the responses was a point of contention. Experts pointed out that climate change effects and solutions are often location-specific, yet the chatbots tended to provide broad, universally applicable answers. This, they suggested, might stem from the nature of the prompts fed to the models. More accurate and tailored responses could potentially be elicited with prompts that include more detailed and specific instructions or context.

    \item \textbf{Expert Insights on Biases and Misinformation in Climate Data Dissemination:} We inquired about the experts' perception of potential biases and misinformation in the chatbot-generated responses on climate-related topics. A notable portion of the experts, about half, identified instances of data exaggeration leading to misinformation. They expressed concerns about the chatbots being trained on datasets with unverified or non-reproducible sources, a significant issue in a field prone to false negatives. Such practices, they cautioned, result in the propagation of unverified and potentially misleading information. Conversely, the other half acknowledged the general adequacy of the information provided by the chatbots but stressed the need for stringent measures to ensure the verifiability of disseminated data. Despite these divergent views, a unanimous concern among all experts was the lack of demographic sensitivity in the chatbots' responses. The experts observed that the chatbots tended to provide uniform answers irrespective of varying demographic contexts. For instance, the response given to a middle-aged African male was identical to that given to a young European female, overlooking the specific vulnerabilities and contexts of different demographic groups. This could mean that the information about the person behind the prompts was not considered much in generating exact answers. One expert poignantly remarked that a teenager and a senior citizen should rather receive relative responses based on their previous knowledge highlighting the need for more nuanced and demographic-aware chatbots.

    \item \textbf{Chatbots as Tools for Promoting Sustainable Behaviors:} We explored the experts' views on the potential of these chatbots in fostering sustainable behaviors and lifestyle changes among users. The response was unanimously positive across the board. The experts acknowledged the importance of reliable information sources but were optimistic about the role of chatbots, especially those specialized in climate change, in influencing user behavior towards sustainability. They concurred that appropriately designed chatbots could effectively encourage users to adopt more environmentally friendly practices. Furthermore, some experts proposed specific features that could enhance the chatbots' capability to advocate for sustainability. These suggestions included personalized advice based on user's lifestyle, interactive guides on reducing carbon footprints, and timely updates on environmental issues and solutions, all tailored to engage users actively in sustainability efforts.
\end{enumerate}

\noindent \textbf{Discussion}: Our panel of climate change experts concurs that, despite the need for considerable improvements in safety and reliability, LLM-backed chatbots hold immense potential for impactful applications. These AI-driven tools are lauded for their capacity to revolutionize the dissemination of crucial information and to promote environmental consciousness among the public. Nonetheless, experts stress the imperative for stringent validation and continuous refinement of these systems to bolster their effectiveness and credibility in information dissemination.\\

\noindent One prominent recommendation from the experts is the strategic curation of training datasets for LLMs, advocating for the inclusion of data primarily from verifiable and expert-endorsed sources. This recommendation arises from their observation that the chatbots in our study often referenced materials indiscriminately, linking to articles that may be obsolete or from publishers lacking official standing in academic research. Moreover, instances of non-functional or potentially deceptive links further highlight the risks associated with unvetted information sources.\\

\noindent The experts also highlighted a critical concern regarding the inherent bias within prompts themselves, suggesting that LLMs may prioritize completing a user's request over ensuring the accuracy and reliability of the provided information. This tendency underscores a fundamental challenge: the need to fine-tune LLMs to discern and prioritize high-quality, trustworthy content in their responses, regardless of the nature of the prompts they receive.

\paragraph{(II.) Experts' Perspectives on Misinformation and Bias in Mental Health:}
In discussions with domain experts, we sought to understand the behavior and potential of the three chatbots in the mental health domain. Our dialogue focused on several key areas:

\begin{enumerate}
    \item \textbf{Impact on Stigma and Awareness:} Our inquiry into experts' perceptions commenced with questions about the potential impact of chatbots on stigma reduction and awareness enhancement in mental health. The response was uniformly positive, with experts recognizing the substantial value of LLM-backed chatbots as a medium for spreading mental health information. Notably, one expert cited current research where AI is leveraged in therapeutic settings to assist psychologists in interpreting clients' emotions and identifying underlying issues more effectively. This ongoing work was paralleled with the evolution of LLMs, suggesting that such advancements could significantly aid in patient support and care.

    \item \textbf{Privacy Concerns:} However, a majority of the experts—five out of seven—raised concerns about privacy when deploying chatbots in mental health contexts. These concerns emerged from a discussion on the ethical implications of using such technologies. The conversation then shifted to the critical role of informed consent in mitigating potential privacy issues associated with the deployment of chatbots. The experts emphasized that user consent should be a cornerstone of any application involving personal and sensitive data.
    
    \item \textbf{Personalized Support and Risks:} The potential of chatbots to offer individualized support was a focal point of our expert discussions. The unanimous viewpoint among the experts was that providing personalized support via chatbots is fraught with risks and is impractical without stringent supervision. They opined that chatbots could serve effectively as conduits of information but should not overstep into realms requiring professional expertise, such as prescribing medication or making clinical diagnoses. One expert succinctly remarked that the chatbots' utility should be confined to information dissemination. This stems from the current trend of utilizing LLMs as an alternative to traditional search engines, leading to a concerning rise in individuals self-diagnosing and self-prescribing without professional medical counsel. Consequently, our experts advocated for a clear demarcation in the chatbots' roles, suggesting that they should act as signposts directing users to qualified health professionals, rather than attempting to replicate the nuanced and critical functions of those experts. This approach is intended to harness the benefits of LLMs in raising awareness while safeguarding against the hazards of misinformed self-care.
    
    \item \textbf{Trust and Confidentiality:} When inquiring whether chatbots could engender a sense of trust and confidentiality for individuals seeking mental health guidance, the experts' feedback was encouragingly positive. A portion of the panel attributed this potential to the chatbots' easy and unrestricted access, positing that the convenience they offer could promote their use as a trustworthy source of support. Others pointed to the broader societal embrace of technology, noting the propensity for individuals to become reliant on apps that simplify complex tasks. They drew parallels with consumer behavior trends, such as the increasing preference for online transactions over in-person interactions and the shift from traditional media consumption to social media engagement. These observations suggest that users may readily confide in chatbots for insights into their mental health challenges, potentially overlooking the inherent risks of substituting professional human interaction with AI-driven advice.

    \item \textbf{Cultural Sensitivity and Inclusivity:} When exploring the inclusivity and cultural sensitivity of these chatbots, our findings revealed diverse opinions among the experts. Two remained neutral, neither confirming nor denying the chatbots' cultural adeptness. However, the majority, five out of seven experts, expressed reservations. They criticized the chatbots for their inability to tailor responses to diverse cultural backgrounds and individual patient details, often providing uniform answers across various demographics. This lack of customization, according to our experts, highlights a need for vigilant oversight in the deployment of these AI systems. Moreover, there was a general consensus that the chatbots' current design predisposes them to a fixed response pattern, suggesting an inherent rigidity that calls for professional human oversight to ensure appropriate and sensitive use.
\end{enumerate}

\noindent \textbf{Discussion:}
In summation, the consensus among the domain experts is that while LLM-backed chatbots hold promise for augmenting professional mental health services, there remains a divergence of opinions regarding their use by individuals for personal mental health needs. The experts commend the chatbots for their ability to encourage users to seek help during crises. An exemplary case is Bard, which provides a helpline number for immediate assistance. Such features exemplify the optimal role of chatbots in mental health care: to provide high-level information and to act as signposts directing users to professional help, rather than serving as standalone therapeutic tools.

\section{Conclusion and Future Work}
\paragraph{Synthesis of Findings:}
Our comprehensive study has shed light on the capabilities and current limitations of LLM-backed chatbots, with a particular focus on ChatGPT, Google Bard, and Bing Chat across two critical domains: Climate Change and Mental Health. The quantitative analysis, as substantiated by our empirical data and illustrated in Table \ref{tab:performance_metrics} and Figure \ref{fig:confusion-matrices-corrected}, demonstrates a proficient performance by the chatbots in True/False classification tasks. Notably, the chatbots showed a higher accuracy rate in the Mental Health domain compared to Climate Change. Furthermore, the normalized similarity index scores, depicted in Figure \ref{fig:similarity_index_scores}, reveal varying degrees of factual adherence, with each chatbot exhibiting unique strengths and weaknesses.

\paragraph{Expert Evaluations:}
Qualitative evaluations by domain experts highlighted the potential of chatbots in enhancing public awareness and reducing stigma associated with mental health. Yet, they also cautioned against unregulated personal use, emphasizing the paramount importance of privacy, ethical considerations, and the need for expert intervention in sensitive scenarios. The experts advocated for chatbots to function as informational gateways, directing users to professional services rather than attempting to replace them.

\paragraph{Recommendations for Further Work:}
Looking ahead, the study identifies several avenues for further research and development. Prominent among these is the need for chatbots to incorporate culturally sensitive and demographically tailored interactions, which currently stands as a significant gap in their design. Additionally, there is a call for the continual refinement of LLM training datasets, ensuring they are derived from verified, expert-approved sources to enhance the quality and reliability of the information provided.

\paragraph{Closing Thoughts:}
In conclusion, while chatbots represent a remarkable technological advancement with the potential to contribute positively to society, their deployment—especially in domains as sensitive as Climate Change and Mental Health—must be approached with meticulous care, ethical standards, and professional oversight. As AI continues to evolve, it is imperative that we, as researchers and developers, keep pace with these advancements, ensuring that we mitigate risks and harness AI's full potential responsibly and ethically. The ultimate goal is to achieve a harmonious integration of AI tools like chatbots into the fabric of societal support systems, augmenting human expertise rather than attempting to supplant it.

{
\bibliography{aaai22}

\begin{thebibliography}{50}
\providecommand{\natexlab}[1]{#1}

\bibitem[{Aremu(2023)}]{7_pandora}
Aremu, T. 2023.
\newblock Unlocking pandora’s box: Unveiling the elusive realm of ai text detection.
\newblock \emph{SSRN Electronic Journal}.

\bibitem[{Baguio, Lu, and Pe{\~n}a(2023)}]{27_climatesentiment}
Baguio, J. D.~S.; Lu, B.~A.; and Pe{\~n}a, C.~F. 2023.
\newblock Text Classification of Climate Change Tweets using Artificial Neural Networks, FastText Word Embeddings, and Latent Dirichlet Allocation.
\newblock \emph{2023 International Conference in Advances in Power, Signal, and Information Technology (APSIT)}, 688--692.

\bibitem[{Banerjee and Lavie(2005)}]{banerjee-lavie-2005-meteor}
Banerjee, S.; and Lavie, A. 2005.
\newblock {METEOR}: An Automatic Metric for {MT} Evaluation with Improved Correlation with Human Judgments.
\newblock In Goldstein, J.; Lavie, A.; Lin, C.-Y.; and Voss, C., eds., \emph{Proceedings of the {ACL} Workshop on Intrinsic and Extrinsic Evaluation Measures for Machine Translation and/or Summarization}, 65--72. Ann Arbor, Michigan: Association for Computational Linguistics.

\bibitem[{Bao et~al.(2023)Bao, Chen, Xiao, Ren, Wu, Zhong, Peng, Huang, and Wei}]{36_mental}
Bao, Z.; Chen, W.; Xiao, S.; Ren, K.; Wu, J.; Zhong, C.; Peng, J.; Huang, X.; and Wei, Z. 2023.
\newblock DISC-MedLLM: Bridging General Large Language Models and Real-World Medical Consultation.
\newblock \emph{ArXiv}, abs/2308.14346.

\bibitem[{Bhardwaj~A and S(2020)}]{46_bias}
Bhardwaj~A, N.~M., Li~J; and S, J. 2020.
\newblock Fairer summarization: Techniques for mitigating bias in story generation.
\newblock \emph{ArXiv}.

\bibitem[{Bommasani et~al.(2022)Bommasani, Hudson, Adeli et~al.}]{2_foundation_models_survey}
Bommasani, R.; Hudson, D.~A.; Adeli, E.; et~al. 2022.
\newblock On the Opportunities and Risks of Foundation Models.
\newblock arXiv:2108.07258.

\bibitem[{Braun and Clarke(2006)}]{45_Braun_Clarke_2006}
Braun, V.; and Clarke, V. 2006.
\newblock Using thematic analysis in psychology.
\newblock \emph{Qualitative Research in Psychology}, 3(2): 77–101.

\bibitem[{Bulian et~al.(2023)Bulian, Sch{\"a}fer, Amini, Lam, Ciaramita, Gaiarin, Huebscher, Buck, Mede, Leippold, and Strauss}]{30.climateassessing}
Bulian, J.; Sch{\"a}fer, M.~S.; Amini, A.; Lam, H.; Ciaramita, M.; Gaiarin, B.; Huebscher, M.~C.; Buck, C.; Mede, N.~G.; Leippold, M.; and Strauss, N. 2023.
\newblock Assessing Large Language Models on Climate Information.
\newblock \emph{ArXiv}, abs/2310.02932.

\bibitem[{Buolamwini and Gebru(2018)}]{11_buolamwini}
Buolamwini, J.; and Gebru, T. 2018.
\newblock Gender Shades: Intersectional Accuracy Disparities in Commercial Gender Classification.
\newblock In Friedler, S.~A.; and Wilson, C., eds., \emph{Proceedings of the 1st Conference on Fairness, Accountability and Transparency}, volume~81 of \emph{Proceedings of Machine Learning Research}, 77--91. PMLR.

\bibitem[{Denecke, Vaaheesan, and Arulnathan(2020)}]{31_mental}
Denecke, K.; Vaaheesan, S.; and Arulnathan, A. 2020.
\newblock A Mental Health Chatbot for Regulating Emotions (SERMO) - Concept and Usability Test.
\newblock \emph{IEEE Transactions on Emerging Topics in Computing}, 9: 1170--1182.

\bibitem[{Diggelmann et~al.(2020)Diggelmann, Boyd-Graber, Bulian, Ciaramita, and Leippold}]{22_climatedata}
Diggelmann, T.; Boyd-Graber, J.~L.; Bulian, J.; Ciaramita, M.; and Leippold, M. 2020.
\newblock CLIMATE-FEVER: A Dataset for Verification of Real-World Climate Claims.
\newblock \emph{ArXiv}, abs/2012.00614.

\bibitem[{Fard, Hasan, and Bell(2022)}]{18_climatchat}
Fard, B.; Hasan, S.~A.; and Bell, J.~E. 2022.
\newblock CliMedBERT: A Pre-trained Language Model for Climate and Health-related Text.
\newblock \emph{ArXiv}, abs/2212.00689.

\bibitem[{Garrido-Merch'an, Gonz'alez-Barthe, and Vaca(2023)}]{20_climatechat}
Garrido-Merch'an, E.~C.; Gonz'alez-Barthe, C.; and Vaca, M.~C. 2023.
\newblock Fine-tuning ClimateBert transformer with ClimaText for the disclosure analysis of climate-related financial risks.
\newblock \emph{ArXiv}, abs/2303.13373.

\bibitem[{Gebru et~al.(2021)Gebru, Morgenstern, Vecchione, Vaughan, Wallach, au2, and Crawford}]{9_gebru}
Gebru, T.; Morgenstern, J.; Vecchione, B.; Vaughan, J.~W.; Wallach, H.; au2, H. D.~I.; and Crawford, K. 2021.
\newblock Datasheets for Datasets.
\newblock arXiv:1803.09010.

\bibitem[{Jain and Padmanaban(2023)}]{29_climatesustainability}
Jain, A.; and Padmanaban, M. 2023.
\newblock Scope 3 emission estimation using large language models.

\bibitem[{Ji et~al.(2021)Ji, Zhang, Ansari, Fu, Tiwari, and Cambria}]{32_mental}
Ji, S.; Zhang, T.; Ansari, L.; Fu, J.; Tiwari, P.; and Cambria, E. 2021.
\newblock MentalBERT: Publicly Available Pretrained Language Models for Mental Healthcare.
\newblock In \emph{International Conference on Language Resources and Evaluation}.

\bibitem[{Kasai et~al.(2023)Kasai, Kasai, Sakaguchi, Yamada, and Radev}]{43_mental}
Kasai, J.; Kasai, Y.; Sakaguchi, K.; Yamada, Y.; and Radev, D.~R. 2023.
\newblock Evaluating GPT-4 and ChatGPT on Japanese Medical Licensing Examinations.
\newblock \emph{ArXiv}, abs/2303.18027.

\bibitem[{Kelley et~al.(2021)Kelley, Yang, Heldreth, Moessner, Sedley, Kramm, Newman, and Woodruff}]{6_perception}
Kelley, P.~G.; Yang, Y.; Heldreth, C.; Moessner, C.; Sedley, A.; Kramm, A.; Newman, D.~T.; and Woodruff, A. 2021.
\newblock Exciting, Useful, Worrying, Futuristic: Public Perception of Artificial Intelligence in 8 Countries.
\newblock 627–637.

\bibitem[{Kraus et~al.(2023)Kraus, Bingler, Leippold, Schimanski, Senni, Stammbach, Vaghefi, and Webersinke}]{21_climatechat}
Kraus, M.; Bingler, J.~A.; Leippold, M.; Schimanski, T.; Senni, C.~C.; Stammbach, D.; Vaghefi, S.~A.; and Webersinke, N. 2023.
\newblock Enhancing Large Language Models with Climate Resources.
\newblock \emph{ArXiv}, abs/2304.00116.

\bibitem[{Krishnan and Anoop(2023)}]{25_climatesentiment}
Krishnan, A.; and Anoop, V.~S. 2023.
\newblock ClimateNLP: Analyzing Public Sentiment Towards Climate Change Using Natural Language Processing.

\bibitem[{Laud et~al.(2023)Laud, Spokoyny, Corringham, and Berg-Kirkpatrick}]{24_climatedata}
Laud, T.~A.; Spokoyny, D.~M.; Corringham, T.~W.; and Berg-Kirkpatrick, T. 2023.
\newblock ClimaBench: A Benchmark Dataset For Climate Change Text Understanding in English.
\newblock \emph{ArXiv}, abs/2301.04253.

\bibitem[{Li et~al.(2023{\natexlab{a}})Li, Gan, Yang, Yang, Li, Wang, and Gao}]{1_foundation_models_multimodal}
Li, C.; Gan, Z.; Yang, Z.; Yang, J.; Li, L.; Wang, L.; and Gao, J. 2023{\natexlab{a}}.
\newblock Multimodal Foundation Models: From Specialists to General-Purpose Assistants.
\newblock arXiv:2309.10020.

\bibitem[{Li(2023)}]{19_climatechat}
Li, Y. 2023.
\newblock Domain Adaptation to Climate Change with Improved BLEU Evaluation Method.

\bibitem[{Li et~al.(2023{\natexlab{b}})Li, Li, Zhang, Dan, Jiang, and Zhang}]{33_mental}
Li, Y.; Li, Z.; Zhang, K.; Dan, R.; Jiang, S.; and Zhang, Y. 2023{\natexlab{b}}.
\newblock ChatDoctor: A Medical Chat Model Fine-Tuned on a Large Language Model Meta-AI (LLaMA) Using Medical Domain Knowledge.
\newblock \emph{Cureus}, 15.

\bibitem[{Liang et~al.(2023)Liang, Yuksekgonul, Mao, Wu, and Zou}]{8_bias}
Liang, W.; Yuksekgonul, M.; Mao, Y.; Wu, E.; and Zou, J. 2023.
\newblock GPT detectors are biased against non-native English writers.
\newblock arXiv:2304.02819.

\bibitem[{Lin(2004)}]{lin-2004-rouge}
Lin, C.-Y. 2004.
\newblock {ROUGE}: A Package for Automatic Evaluation of Summaries.
\newblock In \emph{Text Summarization Branches Out}, 74--81. Barcelona, Spain: Association for Computational Linguistics.

\bibitem[{Liu et~al.(2023{\natexlab{a}})Liu, Zhou, Hua, Chong, Tian, Liu, Wang, You, Guo, Zhu, and Li}]{40_mental}
Liu, J.; Zhou, P.; Hua, Y.; Chong, D.; Tian, Z.; Liu, A.; Wang, H.; You, C.; Guo, Z.; Zhu, L.; and Li, M.~L. 2023{\natexlab{a}}.
\newblock Benchmarking Large Language Models on CMExam - A Comprehensive Chinese Medical Exam Dataset.
\newblock \emph{ArXiv}, abs/2306.03030.

\bibitem[{Liu et~al.(2023{\natexlab{b}})Liu, Li, Cao, Ren, Liao, and Wu}]{35_mental}
Liu, J.~M.; Li, D.; Cao, H.; Ren, T.; Liao, Z.; and Wu, J. 2023{\natexlab{b}}.
\newblock ChatCounselor: A Large Language Models for Mental Health Support.
\newblock \emph{ArXiv}, abs/2309.15461.

\bibitem[{Manathunga and Hettigoda(2023)}]{41_mental}
Manathunga, S.; and Hettigoda, I. 2023.
\newblock Aligning Large Language Models for Clinical Tasks.
\newblock \emph{ArXiv}, abs/2309.02884.

\bibitem[{Mitchell et~al.(2019)Mitchell, Wu, Zaldivar, Barnes, Vasserman, Hutchinson, Spitzer, Raji, and Gebru}]{13_ini}
Mitchell, M.; Wu, S.; Zaldivar, A.; Barnes, P.; Vasserman, L.; Hutchinson, B.; Spitzer, E.; Raji, I.~D.; and Gebru, T. 2019.
\newblock Model Cards for Model Reporting.
\newblock In \emph{Proceedings of the Conference on Fairness, Accountability, and Transparency}, FAT* '19, 220–229. New York, NY, USA: Association for Computing Machinery.
\newblock ISBN 9781450361255.

\bibitem[{Mittelstadt et~al.(2016)Mittelstadt, Allo, Taddeo, Wachter, and Floridi}]{14_brent}
Mittelstadt, B.~D.; Allo, P.; Taddeo, M.; Wachter, S.; and Floridi, L. 2016.
\newblock The ethics of algorithms: Mapping the debate.
\newblock \emph{Big Data \& Society}, 3(2): 2053951716679679.

\bibitem[{Moor et~al.(2023)Moor, Banerjee, Abad, Krumholz, Leskovec, Topol, and Rajpurkar}]{3_foundation_models_medical}
Moor, M.; Banerjee, O.; Abad, Z.~S.; Krumholz, H.~M.; Leskovec, J.; Topol, E.~J.; and Rajpurkar, P. 2023.
\newblock Foundation models for generalist medical artificial intelligence.
\newblock \emph{Nature}, 616(7956): 259–265.

\bibitem[{Ni et~al.(2023)Ni, Bingler, Colesanti-Senni, Kraus, Gostlow, Schimanski, Stammbach, Vaghefi, Wang, Webersinke, Wekhof, Yu, and Leippold}]{17_climatechat}
Ni, J.; Bingler, J.~A.; Colesanti-Senni, C.; Kraus, M.; Gostlow, G.; Schimanski, T.; Stammbach, D.; Vaghefi, S.~A.; Wang, Q.; Webersinke, N.; Wekhof, T.; Yu, T.; and Leippold, M. 2023.
\newblock CHATREPORT: Democratizing Sustainability Disclosure Analysis through LLM-based Tools.
\newblock \emph{ArXiv}, abs/2307.15770.

\bibitem[{Nori et~al.(2023)Nori, King, McKinney, Carignan, and Horvitz}]{39_mental}
Nori, H.; King, N.; McKinney, S.~M.; Carignan, D.; and Horvitz, E. 2023.
\newblock Capabilities of GPT-4 on Medical Challenge Problems.
\newblock \emph{ArXiv}, abs/2303.13375.

\bibitem[{Papineni et~al.(2002)Papineni, Roukos, Ward, and Zhu}]{papineni-etal-2002-bleu}
Papineni, K.; Roukos, S.; Ward, T.; and Zhu, W.-J. 2002.
\newblock {B}leu: a Method for Automatic Evaluation of Machine Translation.
\newblock In Isabelle, P.; Charniak, E.; and Lin, D., eds., \emph{Proceedings of the 40th Annual Meeting of the Association for Computational Linguistics}, 311--318. Philadelphia, Pennsylvania, USA: Association for Computational Linguistics.

\bibitem[{Raghavan et~al.(2019)Raghavan, Barocas, Kleinberg, and Levy}]{5_bias_employment}
Raghavan, M.; Barocas, S.; Kleinberg, J.; and Levy, K. 2019.
\newblock Mitigating bias in algorithmic employment screening: Evaluating claims and practices.
\newblock \emph{SSRN Electronic Journal}.

\bibitem[{Raji et~al.(2020)Raji, Gebru, Mitchell, Buolamwini, Lee, and Denton}]{12_ini}
Raji, I.~D.; Gebru, T.; Mitchell, M.; Buolamwini, J.; Lee, J.; and Denton, E. 2020.
\newblock Saving Face: Investigating the Ethical Concerns of Facial Recognition Auditing.
\newblock In \emph{Proceedings of the AAAI/ACM Conference on AI, Ethics, and Society}, AIES '20, 145–151. New York, NY, USA: Association for Computing Machinery.
\newblock ISBN 9781450371100.

\bibitem[{Ray and Kumar(2023)}]{28_climatesentiment}
Ray, S.; and Kumar, A. M.~S. 2023.
\newblock Prediction and Analysis of Sentiments of Reddit Users towards the Climate Change Crisis.
\newblock \emph{2023 International Conference on Networking and Communications (ICNWC)}, 1--17.

\bibitem[{Rosol et~al.(2023)Rosol, Gasior, Laba, Korzeniewski, and Młyńczak}]{42_mental}
Rosol, M.; Gasior, J.~S.; Laba, J.; Korzeniewski, K.; and Młyńczak, M. 2023.
\newblock Evaluation of the performance of GPT-3.5 and GPT-4 on the Medical Final Examination.
\newblock In \emph{medRxiv}.

\bibitem[{R.S(1998)}]{47_confirmation_bias}
R.S, N. 1998.
\newblock Confirmation bias: A ubiquitous phenomenon in many guises.
\newblock \emph{Review of General Psychology, 2(2), 175}.

\bibitem[{Sham and Mohamed(2022)}]{26_climatesentiment}
Sham, N.~M.; and Mohamed, A.~H. 2022.
\newblock Climate Change Sentiment Analysis Using Lexicon, Machine Learning and Hybrid Approaches.
\newblock \emph{Sustainability}.

\bibitem[{Shneiderman(2020)}]{10_sch}
Shneiderman, B. 2020.
\newblock Bridging the gap between ethics and practice: guidelines for reliable, safe, and trustworthy human-centered AI systems.
\newblock \emph{ACM Transactions on Interactive Intelligent Systems (TiiS)}, 10(4): 1--31.

\bibitem[{Singhal et~al.(2023)Singhal, Tu, Gottweis, Sayres, Wulczyn, Hou, Clark, Pfohl, Cole-Lewis, Neal, Schaekermann, Wang, Amin, Lachgar, Mansfield, Prakash, Green, Dominowska, y~Arcas, Tomavsev, Liu, Wong, Semturs, Mahdavi, Barral, Webster, Corrado, Matias, Azizi, Karthikesalingam, and Natarajan}]{44_mental}
Singhal, K.; Tu, T.; Gottweis, J.; Sayres, R.; Wulczyn, E.; Hou, L.; Clark, K.; Pfohl, S.~R.; Cole-Lewis, H.~J.; Neal, D.; Schaekermann, M.; Wang, A.; Amin, M.; Lachgar, S.; Mansfield, P.~A.; Prakash, S.; Green, B.; Dominowska, E.; y~Arcas, B.~A.; Tomavsev, N.; Liu, Y.; Wong, R.~C.; Semturs, C.; Mahdavi, S.~S.; Barral, J.~K.; Webster, D.~R.; Corrado, G.~S.; Matias, Y.; Azizi, S.; Karthikesalingam, A.; and Natarajan, V. 2023.
\newblock Towards Expert-Level Medical Question Answering with Large Language Models.
\newblock \emph{ArXiv}, abs/2305.09617.

\bibitem[{Spokoyny et~al.(2023)Spokoyny, Laud, Corringham, and Berg-Kirkpatrick}]{23_climatedata}
Spokoyny, D.~M.; Laud, T.~A.; Corringham, T.~W.; and Berg-Kirkpatrick, T. 2023.
\newblock Towards Answering Climate Questionnaires from Unstructured Climate Reports.

\bibitem[{Vaghefi et~al.(2023)Vaghefi, Wang, Muccione, Ni, Kraus, Bingler, Schimanski, Colesanti-Senni, Webersinke, Huggel, and Leippold}]{16_chatclimate}
Vaghefi, S.~A.; Wang, Q.; Muccione, V.; Ni, J.; Kraus, M.; Bingler, J.~A.; Schimanski, T.; Colesanti-Senni, C.; Webersinke, N.; Huggel, C.; and Leippold, M. 2023.
\newblock chatClimate: Grounding Conversational AI in Climate Science.
\newblock \emph{ArXiv}, abs/2304.05510.

\bibitem[{Wang, Zhao, and Petzold(2023)}]{38_mental}
Wang, Y.; Zhao, Y.; and Petzold, L. 2023.
\newblock Are Large Language Models Ready for Healthcare? A Comparative Study on Clinical Language Understanding.
\newblock \emph{ArXiv}, abs/2304.05368.

\bibitem[{Webersinke et~al.(2022)Webersinke, Kraus, Bingler, and Leippold}]{15_webersinke}
Webersinke, N.; Kraus, M.; Bingler, J.~A.; and Leippold, M. 2022.
\newblock ClimateBert: A Pretrained Language Model for Climate-Related Text.
\newblock arXiv:2110.12010.

\bibitem[{Weisz et~al.(2023)Weisz, Muller, He, and Houde}]{4_generative_ai}
Weisz, J.~D.; Muller, M.; He, J.; and Houde, S. 2023.
\newblock Toward General Design Principles for Generative AI Applications.
\newblock arXiv:2301.05578.

\bibitem[{Xu et~al.(2023)Xu, Yao, Dong, Gabriel, Yu, Hendler, Ghassemi, Dey, and Wang}]{34_mental}
Xu, X.; Yao, B.; Dong, Y.; Gabriel, S.; Yu, H.; Hendler, J.; Ghassemi, M.; Dey, A.~K.; and Wang, D. 2023.
\newblock Mental-LLM: Leveraging Large Language Models for Mental Health Prediction via Online Text Data.
\newblock arXiv:2307.14385.

\bibitem[{Yang et~al.(2023)Yang, Ji, Zhang, Xie, and Ananiadou}]{37_mental}
Yang, K.; Ji, S.; Zhang, T.; Xie, Q.; and Ananiadou, S. 2023.
\newblock On the Evaluations of ChatGPT and Emotion-enhanced Prompting for Mental Health Analysis.
\newblock \emph{ArXiv}, abs/2304.03347.

\end{thebibliography}
}
\end{document}